\documentclass[10pt,twocolumn,letterpaper]{article}

\usepackage[pagenumbers]{cvpr} 

\usepackage{graphicx}
\usepackage{amsmath}
\usepackage{amssymb}
\usepackage{booktabs}

\usepackage{epsfig}
\usepackage{graphicx}
\usepackage{amsmath}

\usepackage{amssymb}
\usepackage{makecell}
\usepackage[para,online,flushleft]{threeparttable}
\usepackage{tabularx}
\usepackage{comment}

\makeatletter
\def\widebar{\accentset{{\cc@style\underline{\mskip10mu}}}}
\def\Widebar{\accentset{{\cc@style\underline{\mskip8mu}}}}
\makeatother

\usepackage{bm} 
\usepackage{multirow} 
\usepackage{hhline} 
\usepackage{array} 
\newcolumntype{M}[1]{>{\centering\arraybackslash}m{#1}} 
\usepackage[utf8]{inputenc}
\usepackage[english]{babel}

\usepackage{anyfontsize} 
\usepackage{mathtools} 

\newcommand{\ieno}{\textit{i}.\textit{e}.}
\newcommand{\egno}{\textit{e}.\textit{g}.} 

\newlength{\textfloatsepsave} 
\usepackage{varwidth}
\usepackage{lipsum,etoolbox}
\usepackage{graphics}

\newcolumntype{C}[1]{>{\centering\arraybackslash}m{#1}}
\newcolumntype{R}[1]{>{\raggedleft\arraybackslash}m{#1}}
\newcolumntype{P}[1]{>{\raggedright\arraybackslash}p{#1}}
\newcolumntype{M}[1]{>{\centering\arraybackslash}m{#1}}

\usepackage[pagebackref,breaklinks,colorlinks]{hyperref}

\usepackage[capitalize]{cleveref}
\crefname{section}{Sec.}{Secs.}
\Crefname{section}{Section}{Sections}
\Crefname{table}{Table}{Tables}
\crefname{table}{Tab.}{Tabs.}

\def\cvprPaperID{3054} 
\def\confName{CVPR}
\def\confYear{2022}

\begin{document}
	
	\title{Calibrated Feature Decomposition for Generalizable Person Re-Identification}
	
	\author{Kecheng Zheng, Jiawei Liu, Wei Wu, Liang Li, Zheng-jun Zha\\
		University of Science and Technology of China\\
		{\tt\small zkcys001@mail.ustc.edu.cn}
	}
	\maketitle
	
	\begin{abstract}
		Existing disentangled-based methods for generalizable person re-identification aim at directly disentangling person representations into domain-relevant interference and identity-relevant feature. However, they ignore that some crucial characteristics are stubbornly entwined in both the domain-relevant interference and identity-relevant feature, which are intractable to decompose in an unsupervised manner. In this paper, we propose a simple yet effective Calibrated Feature Decomposition (CFD) module that focuses on improving the generalization capacity for person re-identification through a more judicious feature decomposition and reinforcement strategy. Specifically, a calibrated-and-standardized Batch normalization (CSBN) is designed to learn calibrated person representation by jointly exploring intra-domain calibration and inter-domain standardization of multi-source domain features. CSBN restricts instance-level inconsistency of feature distribution for each domain and captures intrinsic domain-level specific statistics. The calibrated person representation is subtly decomposed into the identity-relevant feature, domain feature, and the remaining entangled one. For enhancing the generalization ability and ensuring high discrimination of the identity-relevant feature, a calibrated instance normalization (CIN) is introduced to enforce discriminative id-relevant information, and filter out id-irrelevant information, and meanwhile the rich complementary clues from the remaining entangled feature are further employed to strengthen it. Extensive experiments demonstrate the strong generalization capability of our framework. Our models empowered by CFD modules significantly outperform the state-of-the-art domain generalization approaches on multiple widely-used benchmarks. Code will be made public: \url{https://github.com/zkcys001/CFD}.
		
		
	\end{abstract}
	
	\section{Introduction}
	
	\begin{figure}[!t]
		\centering
		\includegraphics[width=0.44\textwidth]{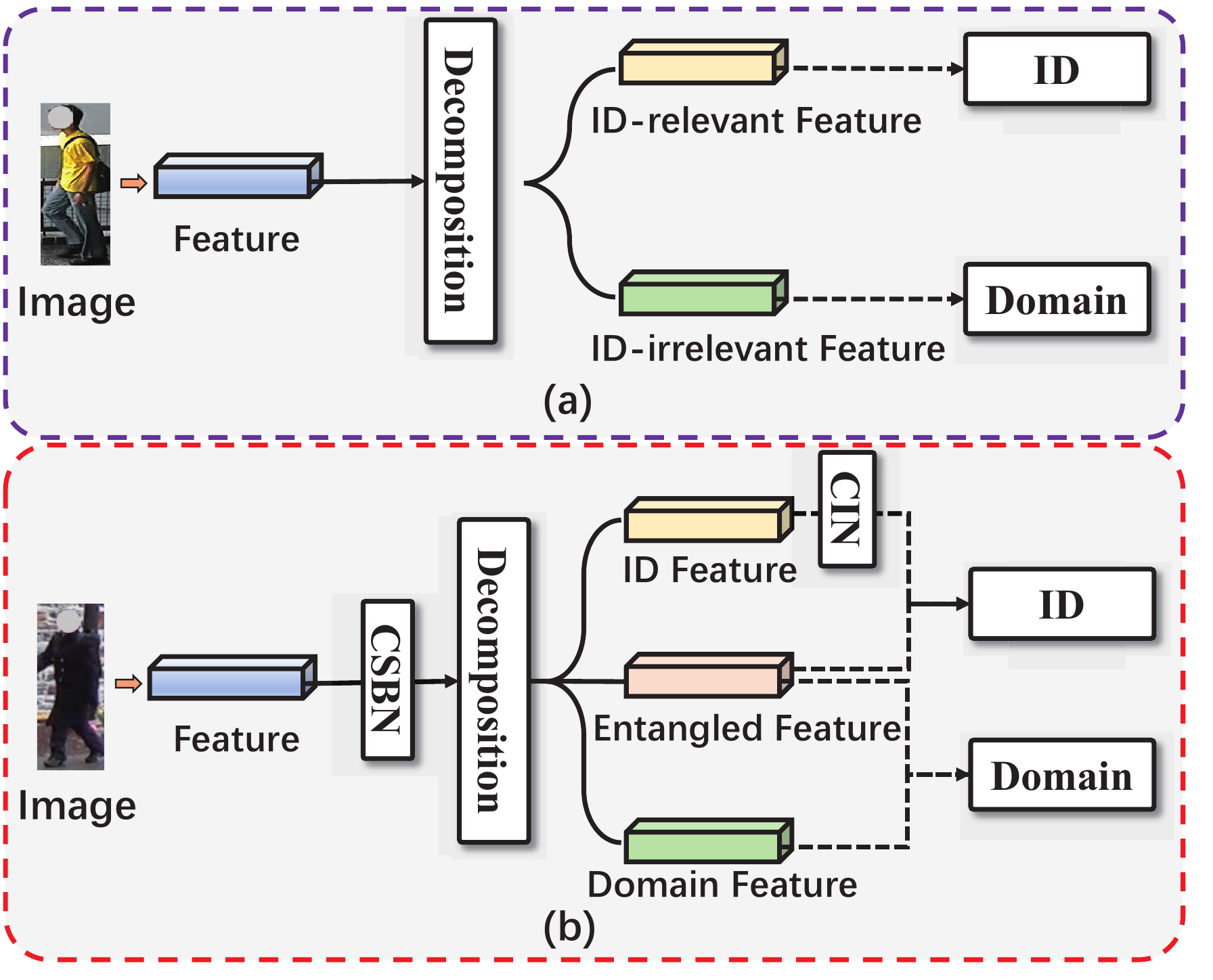}
		\caption{Illustration of motivation and our idea. (a) Conventional disentangled-based ReID method. They often make the strong assumption that the person representation can be decomposed into the id-relevant feature and id-irrelevant feature, and they also ignore to adopt a suitable reinforcement strategy as the guidance of feature decomposition; 
			(b) Our proposed CDM module proposes a more judicious decomposition strategy to decompose person representation into the identity-relevant feature, domain feature, and the remaining entangled one. Meanwhile, we also adopt the CSBN and CIN to facilitate feature decomposition.}
		\label{fig:intro}
		\vspace{-10pt}
	\end{figure}

	Person Re-identification (re-id)\cite{zhang2016learning,su2017pose,zhao2017spindle,zheng2021group} has achieved remarkable success under the transductive domain adaptive setting with the assumption that training and testing data are easily collected. However, this assumption is often violated in practical applications, as new domain data are not always accessible due to privacy issues and expensive labeling costs, which causes dramatic performance degradation on the unseen target domain. For example, the model trained on the data collected in the daytime domain performs poorly on unseen nighttime domains. Therefore, recent efforts have been devoted to Domain Generalizable Person Re-Identification (dubbed as DG-ReID) problem that aims to learn a general domain-agnostic model, which can generalize well to unseen target domains.
	
	Although plenty of domain generalization methods~\cite{chattopadhyay2020learning,huang2020self} (\ieno, data manipulation, representation learning and learning strategy) are proposed, they have been often confined to the situation that the source and target domains share the same label space with a fixed amount of classes. In contrast, domain generalization for person Re-ID is an open-set retrieval task, having different and variable numbers of identities between source and target domains. Therefore, it is difficult to achieve satisfying generalization capability when the existing DG approaches are directly applied to person Re-ID.

	To address this problem, several tailored DG-ReID methods~\cite{zhao2020learning, choi2020meta,song2019generalizable,dai2021generalizable,seo2020learning,zhuang2020rethinking,mancini2018best,he2021semi} have been proposed, which can be mainly divided into three categories: 
	meta-learning based model, ensemble learning based model and disentanglement based model. 
	Ensemble learning based methods~\cite{dai2021generalizable,seo2020learning,zhuang2020rethinking,mancini2018best} assemble multiple domain specific models like experts or classifiers to enhance the generalization ability of the overall network. 
	Meta-learning based models~\cite{zhao2020learning, choi2020meta} combine a model-agnostic meta learning with the normalization technology to mimic real train-test domain shift for improving the generalization of the model. Normalization technique plays an important role in domain generalization task, because it can eliminate id-irrelevant discrepancy to restorate helpful discriminative information. Nevertheless, these methods fail to deeply analyze the impact of different types of normalization techniques placed in different positions on the corresponding features. Meanwhile, these two kinds of methods significantly enlarge model complexity as the number of source domains increases during training.
	
	Compared to the two types of aforementioned methods, as illustrated in Fig.~\ref{fig:intro} (a), the disentanglement based model is a more straightforward and efficient approach for DG-ReID~\cite{jin2020style,eom2019learning,zhang2021disentanglement,zou2020joint,zhang2021learning}. It aims to directly disentangle identity-irrelevant interference and domain-invariant feature from the learned representations, and employ the domain-invariant feature to improve the generalization. However, these methods do not jointly consider the instance-level inconsistency of feature distribution within each domain and intrinsic domain-level specific statistics~\cite{gao2021representative}, which undermine the inherent relevance of unseen target domain with respect to source domains, resulting in unsatisfying generalization capability. Such instance-level inconsistency~\cite{choi2020meta,huang2017arbitrary,gao2021representative} and domain-specific characteristics~\cite{chang2019domain,seo2020learning,d2018domain,piratla2020efficient} can provide adequately discriminative and meaningful information, and guide the model to enable better feature decomposition from the perspective of improving the model generalization. More importantly, these disentanglement based methods ignore that some crucial characteristics are stubbornly entwined in both the domain-relevant interference and identity-relevant feature, which are intractable to decompose in an unsupervised manner. The object of this task is to improve the generalization of the model rather than to perform a complete feature disentanglement. Thus, it is more judicious to decompose the person representation into the identity-relevant feature, domain feature, and the remaining entangled one, and apply suitable normalization techniques to the corresponding features for improving model generalization. This judicious feature decomposition with a suitable reinforcement strategy is the key to the success of DG-ReID.
	
	In this architecture, we propose a simple yet effective Calibrated Feature Decomposition module that focuses on improving the generalization capacity for person re-identification through a more judicious feature decomposition and reinforcement strategy.
	Specifically, in the CFD module, we first employ a CSBN to learn calibrated person representation by jointly exploring intra-domain calibration and inter-domain standardization of multi-source domain features. CSBN restricts instance-level inconsistency of feature distribution within each domain and captures intrinsic domain-level specific statistics. Then we utilize a more judicious feature decomposition to subtly decompose the above calibrated person representation into identity-relevant feature, domain feature, and the remaining entangled one. For enhancing the generalization ability and ensuring high discrimination of identity-relevant feature, CIN is introduced to enforce discriminative id-relevant information and filter out id-irrelevant information. However, such a process inevitably ignores the entangled id-related information, we consider employing the rich complementary clues from the remaining entangled feature to strengthen the id-relevant feature. In addition, domain loss is employed to ensure that id-irrelevant domain information does not flow into the subsequent network. Extensive experiments demonstrate the strong generalization capability of our framework. Our models empowered by the CFD modules significantly outperform the state-of-the-art domain generalization approaches on multiple widely-used person ReID benchmarks.

	The main contributions of this paper are as following:
	
	\begin{itemize}

		\item We propose a domain generalizable person ReID method that generalizes well on unseen domains. Particularly, we design a Calibrated Feature Decomposition module, which is simple yet effective and can be used as a plug-and-play module for existing ReID architectures to enhance their generalization capability.  
		
		\item Our CDM module has a more judicious feature decomposition for enhancement of model generalization capabilities, which subtly decompose the person representation into the identity-relevant feature, domain feature, and the remaining entangled one. 
		
		\item We propose two simple yet effective normalization-based reinforcement strategies that complement each other to improve the generalization and robustness of the model. 
		CSBN is adopted to learn calibrated person representation by jointly exploring intra-domain calibration and inter-domain standardization of multi-source domain features;
		CIN is used to eliminate some style discrepancies followed by entangled identity-relevant information restitution.

	\end{itemize}

	\section{Related Work}

	\subsection{Domain Generalizable Re-IDentification} 
	Generalization capability to unseen domains is crucial for person re-id models when deploying to practical applications. 
	To address this problem, several tailored DG-ReID methods~\cite{zhao2020learning, choi2020meta,song2019generalizable,dai2021generalizable,seo2020learning,zhuang2020rethinking,mancini2018best,lin2020domain} have been proposed, which can be mainly divided into three categories: 
	meta-learning based model, ensemble learning based model and disentanglement based model. 
	Due to the success of disentangled learning, the DG-ReID methods based on disentangled learning~\cite{jin2020style,eom2019learning,zhang2021disentanglement,zou2020joint} improve the model generalization ability by disentangling person representations into identity-irrelevant interference and id-invariant feature. 
	Specifically, Eom \textit{et al.} \cite{eom2019learning} propose to disentangle identity-related and -unrelated features from person images and adopt identity shuffle GAN to enhance the person representation.
	Zhang \textit{et al.} \cite{zhang2021disentanglement} propose a Disentanglement-based Cross-Domain Feature Augmentation (DCDFA) strategy to generate virtual samples in the feature space by adding disentangled domain-specific enhancements upon disentangled domain-shared identity bases.
	Zhang \textit{et al.}~\cite{zhang2021learning} construct a structural causal model (SCM) to disentangle the person representation into identity-specific factors and domain-specific factors.
	Jin \textit{et al.} \cite{jin2020style} design the Style Normalization and Restitution (SNR) module based on instance normalization and feature distillation. 
	However, these disentanglement based models ignore that some crucial characteristics are stubbornly entwined in both the domain-relevant interference and identity-relevant feature, which are intractable to decompose in an unsupervised manner.
	The goal of DG-ReID is to improve the generalization of the model rather than to perform a complete feature disentanglement. 
	Thus, a more judicious feature decomposition with a suitable reinforcement strategy is important for the success of DG-ReID.

	\subsection{Normalization in DA and DG}
	Normalization techniques in deep neural networks are designed for regularizing trained models and improving their generalization performance. Recently, several methods on domain adaptation (DA) and domain generalization (DG) discovered the relationship between domain gap and normalization operation. For example, Jin \textit{et al.}~\cite{jin2021style} proposed a Style Normalization and Restitution module, which utilizes the Instance Normalization (IN) layers to filter out style variations and compensates for the identity-relevant features discarded by IN layers. Seo \textit{et al.}~\cite{seo2020learning} proposed to leverage instance normalization to optimize the trade-off between cross-category variance and domain invariance, which is desirable for domain generalization in unseen domains. Zhuang \textit{et al.}~\cite{zhuang2020rethinking} proposed camera-based batch normalization (CBN) to force the images of all cameras to fall onto the same subspace and to shrink the distribution gap between any camera pair.

	\section{Approach}

\begin{figure*}[!t]
		\centering
		\includegraphics[width=0.99\textwidth]{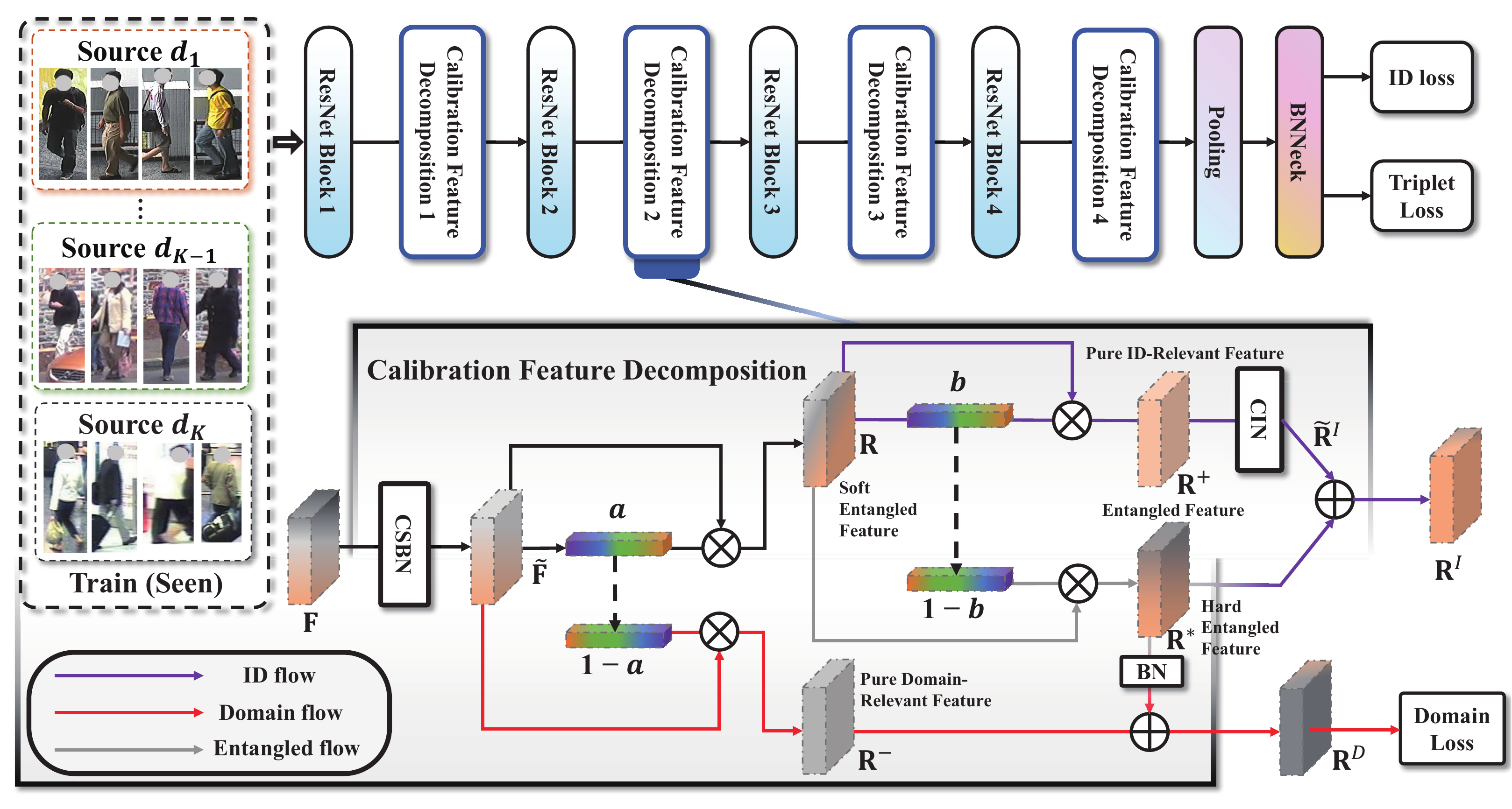}
		\caption{Overall flowchart. (a) Our generalizable person ReID network with the proposed CDM module being plugged in after some convolutional blocks. Here, we use ResNet-50 as our backbone for illustration.
			(b) The proposed CFD module. The calibrated-and-standardized Batch normalization is designed to learn calibrated person representation by jointly exploring intra-domain calibration and inter-domain standardization of multi-source domain features. A calibrated instance normalization is introduced to enforce discriminative id-relevant information and filter out id-irrelevant information, and meanwhile the rich complementary clues from the remaining entangled feature are further employed to strengthen it.}
		\label{fig:framework}
		\vspace{-6pt}
	\end{figure*}

	\noindent\textbf{Overview.} 
	Fig.~\ref{fig:framework} shows the overall architecture of our framework. In this architecture, we propose a simple yet effective Calibrated Feature Decomposition (CFD) module that focuses on improving the generalization capacity for person re-identification through a more judicious feature decomposition and reinforcement strategy.
	Taking the widely used ResNet-50~\cite{he2016deep} in ReID network as a backbone, CFD module as a plug-and-play technology can be easily added after each convolutional block and is extremely easy to implement.
	Specifically, in the CFD module, we first employ a CSBN to learn calibrated person representation by jointly exploring intra-domain calibration and inter-domain standardization of multi-source domain features. 
	CSBN restricts instance-level inconsistency of feature distribution for each domain and captures intrinsic domain-level specific statistics. 
	Then we utilize a more judicious feature decomposition to subtly decompose the above calibrated person representation into the identity-relevant feature, domain feature, and the remaining entangled one. 
	For enhancing the generalization ability and ensuring high discrimination of identity-relevant feature, CIN is introduced to enforce discriminative id-relevant information and filter out id-irrelevant information.
	However, such a process inevitably ignores the entangled id-related information, we consider employing the rich complementary clues from the remaining entangled feature to strengthen id-relevant information. In addition, a domain loss is used to ensure that id-irrelevant domain information does not flow into the subsequent network.
	
	We aim at constructing an efficient, generalizable, and robust person ReID framework for the domain generalizable person re-identification task.

	\subsection{Calibrated Feature Decomposition (CFD)}
	\label{sec:cfd}
	
	In this section, we firstly introduce the novel feature decomposition mechanism for DG re-ID. For a CFD module, we denote the input that is from the ResNet Block by $\mathbf F \in \mathbb{R}^{h\times w \times c}$, and the outputs by $\mathbf{R}^{I},\mathbf{R}^{D} \in \mathbb{R}^{h\times w \times c}$, where $h,w,c$ denote the height, width, and number of channels, respectively.
	
	Given $\mathbf F$ from one ResNet Block, we firstly employ a CSBN to learn calibrated person representation $ \widetilde{\mathbf F}$ by jointly exploring intra-domain calibration and inter-domain standardization of multi-source domain features. 
	Then, this calibrated person representation $ \widetilde{\mathbf F} $ is decomposed into two parts: soft entangled feature $\mathbf R \in \mathbb{R}^{h\times w\times c}$ and pure domain-relevant feature $\mathbf R^- \in \mathbb{R}^{h\times w\times c}$ (\ieno, this feature contains the domain information that is irrelevant to id information). 
	To adequately filter out this pure domain-relevant feature $\mathbf R^-$, following the previous disentanglement-based ReID methods~\cite{jin2020style,zhang2021disentanglement}, we utilize learnable channel attention vector $\mathbf{a}=[a_1, a_2, \cdots, a_c] \in \mathbb{R}^c$ for soft feature decomposition:
	\begin{equation}
		\begin{aligned}
			\mathbf R(:,:,k) = & a_k \widetilde{\mathbf F}(:,:,k), \\ \mathbf R^-(:,:,k) = & (1 - a_k)  \widetilde{\mathbf F}(:,:,k),
		\end{aligned}
		\label{eq:seperation}
	\end{equation}
	where $\mathbf R \in \mathbb{R}^{h\times w \times c}$ refers to the soft entangled feature, consisting of the pure identity-relevant feature and hard entangled feature. 
	We expect that the channel attention vector $\mathbf{a}$ to enable the adaptive distillation of the domain-relevant features for filtering out the id-unrelated feature, and derive it by channel attention as:
	\begin{equation}
		\begin{aligned}
			\mathbf{a} = g(\widetilde{\mathbf F}) = \sigma({\rm \mathbf W_2}\cdot\delta({\rm \mathbf W_1} \cdot pool(\widetilde{\mathbf F}))),
		\end{aligned}
		\label{eq:se}
	\end{equation}
	where $pool(\cdot)$ refers to a global average pooling layer followed by two fully-connected layers that are parameterized by ${{\rm \mathbf  W_2}} \in \mathbb{R}^{(c/r) \times c}$ and ${{\rm \mathbf W_1}} \in \mathbb{R}^{c \times  (c/r)}$. ReLU $\delta(\cdot)$ and sigmoid $\sigma(\cdot)$ are utilized as the activation function, respectively. In addition, a dimension reduction ratio $r$ is used to reduce the number of parameter and is set to 16.
	
	In a same way, for distilling the better id-relevant information, we also adopt the channel attention to decompose the soft entangled feature $\mathbf R$ into hard entangled feature $\mathbf R^{*} \in \mathbb{R}^{h\times w\times c}$ and pure id-relevant feature $\mathbf R^+ \in \mathbb{R}^{h\times w\times c}$:
	\begin{equation}
		\begin{aligned}
			\mathbf R^*(:,:,k) = & b_k \mathbf R(:,:,k), \\ \mathbf R^+(:,:,k) = & (1 - b_k)  \mathbf R(:,:,k),
		\end{aligned}
		\label{eq:seperation2}
	\end{equation}
	
	To enforce discriminative id-relevant information and filter out id-irrelevant information, a CIN is employed to the $\mathbf R^+$ to obtain the calibrated pure id-relevant feature $\widetilde{\mathbf R}^+$, and meanwhile the rich complementary clues from the remaining entangled feature $\mathbf R^{*}$ is further employed to strengthen $\widetilde{\mathbf R}^+$.
	Moreover, to ensure that id-irrelevant domain information does not flow into the main branch as much as possible, we also add hard entangled feature $\mathbf R^{*}$ into pure domain-relevant feature ${\mathbf R}^-$, and use the domain classification loss to guide the output feature $\mathbf R^D$:
	
	\begin{equation}
		\vspace{-3pt}
		\begin{aligned}
			\mathbf R^{I} = & \mathbf R^{+}+\mathbf R^*, \\ \mathbf R^D = & \mathbf R^{-}+\mathbf R^*,
		\end{aligned}
		\label{eq:fusion1}
		\vspace{-2pt}
	\end{equation}
	where $\mathbf R^{I} $ refers to the discriminative identity-relevant feature, and $\mathbf R^{D} $ refers to the contributive domain-relevant feature. A few factors of person representations simultaneously are twined in both the domain-relevant interference and identity-relevant feature, which are intractable to decompose. Our goal is to obtain a more generalizable pedestrian feature rather than to have the feature decoupled. So, it is more important to filter out the id-irrelevant feature, and enhance the discrimination of id-related feature, where this strategy is more suitable for DG re-ID.

	\begin{figure}[!t]
		\centering
		\includegraphics[width=0.40\textwidth]{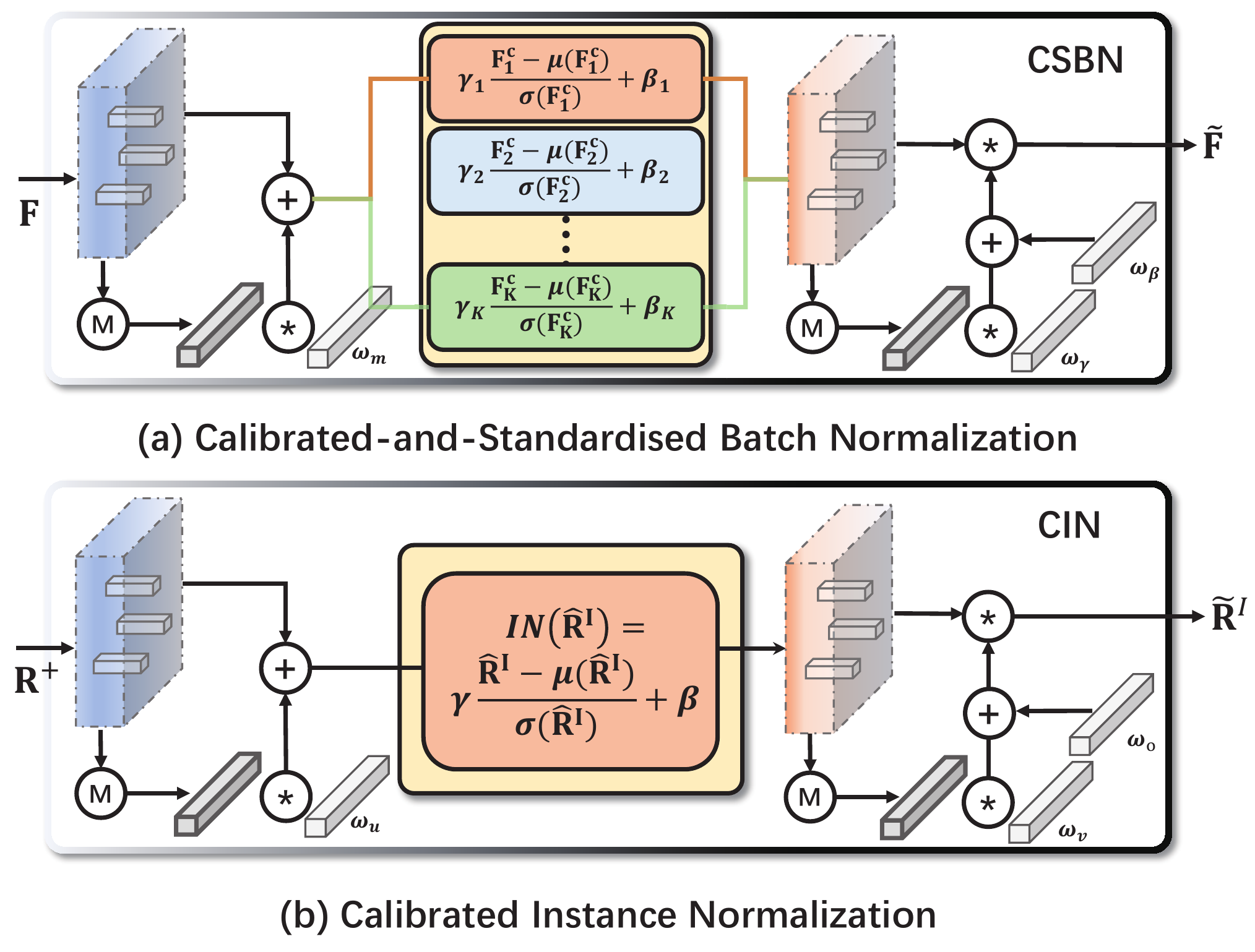}
		\caption{The details of the calibrated-and-standardized batch normalization and the calibrated instance normalization.}
		\label{fig:norm}
		\vspace{-10pt}
	\end{figure}
	
	\subsection{Calibrated Normalization.} 
	
	\noindent\textbf{Calibrated-standardized Batch Normalization.} 
	Previous Batch Normalization (BN), Instance Normalization (IN), or other normalization technologies do not jointly consider intra-domain calibration and inter-domain standardization of multi-source domain features. However, the instance-level inconsistency of feature distribution for each domain and intrinsic domain-level specific statistics are beneficial for the generalization ability of model~\cite{d2018domain,gao2021representative}.
	Thus, to enable the intra-domain calibration, we add a centering calibration scheme driven by instance statistics, where the centering calibration is added before BN layer. Given input feature $\mathbf F$, the centering calibration of features is written as follows:
	\begin{equation}
		\begin{aligned}
			\mathbf F^{c} = \mathbf F+\mathbf{\omega}_m \odot pool(\mathbf F)
		\end{aligned}
	\end{equation}
	where $pool(\cdot)$ refers to a max average pooling layer and $\omega_m$ is the shared learnable weight vector between domains. $\odot$ refers to the dot product, which broadcasts two features to the same dimension size and then conducts dot product.
	
	Because the distribution of different domains varies widely, using a regular BN upon all source domains would result in confusing statistical information. And sharing parameters across multiple domains is inappropriate due to the existence of a domain gap. Inspired by~\cite{d2018domain}, we use domain-specific BN to enable inter-domain standardization of multi-source domain features. When testing, we adopt the mean path of BN parameters to extract the unseen domain feature, which enables better generalization ability. Specifically, let $\mathbf F^{c} = (\mathbf F_{1}^{c}, \mathbf F_{2}^{c}, \ldots, \mathbf F_{K}^{c})$ denotes feature maps of the centering calibrated representation including K domains, CSBN module can be written as:
	
	\begin{equation}
		\begin{aligned}
			\widetilde{\mathbf F}_k =& {\rm {CSBN}}(\mathbf R^{I}) \\&= (\mathbf \gamma_d \frac{\mathbf F_{k}^{c}-\mu(\mathbf F_{d}^{c})}{\sigma(\mathbf F_{k}^{c})} + \mathbf \beta_d)\cdot Sigmoid(\mathbf \omega_{\gamma} \odot \mathbf F_{k}^{c} + \mathbf \omega_{\beta}),
		\end{aligned}
	\end{equation}
	where $\widetilde{\mathbf F} = (\widetilde{\mathbf F}_1, \widetilde{\mathbf F}_2, \ldots, \widetilde{\mathbf F}_K)$, $\widetilde{\mathbf F}_k$ denotes the calibrated person representation of k-th domain, $\mathbf \omega_{\gamma}, \mathbf \omega_{\beta}$ are shared learnable weight vectors between domains, $\mathbf \beta_d, \mathbf \beta_d$ are individual learnable weight vectors in each domain, and $Sigmoid(\cdot)$ means regular sigmoid activation function.
	The calibrated-and-standardized Batch normalization in CFD jointly explores intra-domain calibration and inter-domain standardization of multi-source domain features to perform calibrated feature normalization. Specifically, we utilize the $\mathbf \omega_m$, $\mathbf \omega_{\gamma}$ and $\mathbf \omega_{\beta}$ are shared learnable weight vectors between domains to perform intra-domain calibration. Then, the domain-specific individual parameters (\ieno,  $\mathbf \beta_d, \mathbf \beta_d$) are adopted for the inter-domain standardization.

	\begin{table*}[tb]
		\caption{Performance (\%) comparison with the state-of-the-art methods under Protocol-1. ``MS'' refers to the multiple training datasets, \ieno, Market-1501 (M), DukeMTMC-reID (D), CUHK-SYSU (CS), CUHK03 (C3) and CUHK02 (C2). We mark the results of the best results by \textbf{bold} text.}
		\centering
		\vspace{-0.20cm}
		\resizebox{\linewidth}{!}{
			\begin{tabular}{c|c|c||ccc||ccc||ccc||ccc||ccc}
				\Xhline{3\arrayrulewidth}
				\multirow{2}{*}{Method} & \multirow{2}{*}{Type} & \multirow{2}{*}{Source} & \multicolumn{3}{c||}{Target: VIPeR (V)} & \multicolumn{3}{c||}{Target: PRID (P)} & \multicolumn{3}{c||}{Target: GRID (G)} & \multicolumn{3}{c||}{Target: iLIDS (I)} & \multicolumn{3}{c}{Mean of V,P,G,I}\\ \cline{4-18} 
				&&& R1 & R5  & \emph{m}AP & R1 & R5  & \emph{m}AP & R1 & R5  & \emph{m}AP & R1 & R5 & \emph{m}AP & R1 & R5 & \emph{m}AP \\ \hline
				CDEL~\cite{lin2020domain}                       & DG & M+C2+C3+CS+D     & 38.5 &  -   &   -  & 57.6 &   -  &   -  & 33.0 &   -   &   -  & 62.3 &   -   &   -   & 47.9 & - & -\\
				DIMN~\cite{song2019generalizable}               & DG & M+C2+C3+CS+D     & 51.2 & 70.2 & 60.1 & 39.2 & 67.0 & 52.0 & 29.3 & 53.3  & 41.1 & 70.2 & 89.7  & 78.4  & 47.5 & 70.1 & 57.9 \\
				AugMining~\cite{tamura2019augmented}            & DG & M+C2+C3+CS+D     & 49.8 & 70.8 &  -   & 34.3 & 56.2 &  -   & 46.6 & 67.5  &  -   & 76.3 & 93.0  & -     & 51.8 & 71.9 & - \\
				DualNorm~\cite{jia2019frustratingly}            & DG & M+C2+C3+CS+D     & 53.9 & 62.5 & 58.0 & 60.4 & 73.6 & 64.9 & 41.4 & 47.4  & 45.7 & 74.8 & 82.0  & 78.5  & 57.6 & 66.4 & 61.8\\
				DDAN~\cite{chen2020dual}                        & DG & M+C2+C3+CS+D     & 56.5 & 65.6 & 60.8 & 62.9 & 74.2 & 67.5 & 46.2 & 55.4  & 50.9 & {78.0} & 85.7  & 81.2  & 60.9 & 70.2 & 65.1\\
				RaMoE~\cite{dai2021generalizable}               & DG & M+C2+C3+CS+D     & 56.6 & -    & 64.6 & 57.7 & -    & 67.3 & 46.8 & -     & 54.2 & {85.0} & -     & {90.2}  & 61.5 & - & 69.1\\
				SNR~\cite{jin2020style}                         & DG & M+MT+C3+D        &{55.1}& -    & 65.0 & 49.0 & -    & 60.0 & 30.4 & -  & 41.3 & {91.9} & -  & {87.0}  & 56.6 & - &63.3 \\
				SNR~\cite{jin2020style}                         & DG & M+C2+C3+CS+D     &49.2  &-     & 58.0 & 47.3 &-     & 60.4 & 39.4 & - & 49.0 & 77.3 & - & 84.0 & 53.3  &- & 62.9 \\ 
				CBN~\cite{zhuang2020rethinking}                 & DG & M+C2+C3+CS+D     &49.0  & 63.4 & 59.2 & 61.3 & 73.8 & 65.7 & 43.3 & 48.4& 47.8& 75.3& 84.6& 79.4& 57.2& 67.6& 63.0 \\ 
				Person30K~\cite{bai2021person30k}               & DG & M+C2+C3+CS+D     &53.9  & -    & 60.4 & 60.6 & -    & 68.4 & 50.9 & - &56.6& 79.3& - &83.9& 61.1& - &67.3\\
				DIR-ReID~\cite{zhang2021learning}               & DG & M+C2+C3+CS+D     &{58.3}& 66.9 & 62.9 & 71.1 & 82.4 & 75.6 & 47.8 & 51.1  & 52.1 & 74.4 & 83.1  & 78.6  & 62.9 & 70.8 & 67.3\\
				MetaBIN~\cite{choi2020meta}                     & DG & M+C2+C3+CS+D     & 56.2 &{76.7}&{66.0}& {72.5} & {88.2}  & {79.8} & {49.7} & {67.6}  & {58.1} & {79.7} & {93.3}  & {85.5}  & {64.5} & {81.5} & {72.4}\\
				\midrule
				QAConv$_{50}$~\cite{liao2020interpretable}      & DG & M+C2+C3+CS*      & 57.0&- &66.3 & 52.3&- &62.2 &48.6 &- &57.4 & 75.0& -&81.9 &58.2 &- &67.0 \\
				M$^3$L~\cite{zhao2021learning}                  & DG & M+C2+C3+CS*      & {60.8}& {75.6}  & {68.2} & {55.0} & {79.0}  & {65.3} & {40.0} & {64.0}  & {50.5} & {65.0} & {83.3}  & {74.3}  & {55.2} & {75.5} & {64.6}\\
				MetaBIN~\cite{choi2020meta}                     & DG & M+C2+C3+CS*      & 55.9&- &64.3 & {61.2}&- &{70.8} &50.2 &- &57.9 & 74.7& -&82.7 &{60.5} &- &{68.9}\\
				\textbf{CFD (Ours)}                             & DG & M+C2+C3+CS*      & \textbf{63.3} & \textbf{85.4} & \textbf{72.9} & \textbf{73.0} & \textbf{86.0} & \textbf{79.9} & \textbf{51.0} & \textbf{72.0} & \textbf{60.6} & \textbf{76.7} & \textbf{88.3} & \textbf{82.8}  & \textbf{66.0} & \textbf{82.9} & \textbf{76.4} \\ \Xhline{3\arrayrulewidth}
			\end{tabular}
		}    
		
		\label{table:result1}
		\begin{tablenotes}
			\tiny
			\item[*] We did not include DukeMTMC-reID~\cite{zheng2017unlabeled} in the training domains since this dataset has been discredited by the creators. Although we used one less dataset, our method still has the best average performance on these four small datasets.
		\end{tablenotes}
	\vspace{-.15in}
	\end{table*}

	\begin{table}[h]
		\centering
		\caption{Different evaluation settings of Protocol-1, Protocol-2 and Protocol-3.}
		\vspace{-0.30cm}
		\fontsize{8.5pt}{5.5pt}\selectfont
		\resizebox{\linewidth}{!}{
			\begin{tabular}{c|c|c}
				\toprule
				Setting                       & Training Data & Testing Data \\
				\toprule
				Protocol-1                     & Com-(M+C2+C3+CS) & PRID, GRID, VIPeR, iLIDs \\ 
				\midrule
				\multirow{3}{*}{Protocol-2} & Com-(CS+C3+MT) & M \\ 
				& Com-(M+CS+MT) & C3 \\ 
				& Com-(M+CS+C3) & MT \\ 
				\midrule
				\multirow{4}{*}{Protocol-3} & CS+C3+MT & M \\ 
				& M+CS+MT & C3 \\ 
				& M+CS+C3 & MT \\
				\bottomrule
			\end{tabular}%
		}
		\vspace{-0.50cm}
		\label{table:Protocol}%
	\end{table}%

	\noindent\textbf{Calibrated Instance Normalization} In previous methods~\cite{ulyanov2017improved,dumoulin2016learned,huang2017arbitrary}, they try to reduce the domain discrepancy on the input features by performing IN as:
	\begin{equation}
		\begin{aligned}
			\widetilde{\mathbf R}^{I} = {\rm {IN}}(\mathbf R^{I}) = \mathbf \gamma  (\frac{\mathbf R^{I}-\mu(\mathbf R^{I})}{\sigma(\mathbf R^{I})}) + \mathbf \beta,
		\end{aligned}
	\end{equation}
	where $\mu(\cdot)$ and $\sigma(\cdot)$ denote the mean and standard deviation computed across spatial dimensions independently for each channel and each \emph{sample/instance}, $\mathbf \gamma$, $\mathbf \beta$ $\in \mathbb{R}^c$ are parameters learned from data. 
	IN could filter out some instance-specific style information from the content. But the correct placement of IN is also not considered in the previous decoupling method~\cite{jin2020style,zhang2016learning,eom2019learning}. Because of the nature of IN~\cite{ulyanov2017improved,dumoulin2016learned,huang2017arbitrary}, we put it in the pure ID-related features to better perceive the ID-relevant information from entangled feature. In addition, for enhancing the generalization ability and ensuring high discrimination of identity-relevant feature, we introduce a calibrated operation on the IN layer to enforce discriminative id-relevant information and filter out id-irrelevant information:
	\begin{equation}
		\begin{aligned}
			\widetilde{\mathbf R}^{I} =& {\rm {CIN}}(\mathbf R^{I}) \\ =& (\mathbf \gamma \frac{\hat{\mathbf R}^{I}-\mu(\mathbf R^{I})}{\sigma(\mathbf R^{I})} + \mathbf \beta)\cdot Sigmoid(\mathbf \omega_v \odot \hat{\mathbf R^{I}}+\omega_o), \\ \hat{\mathbf R}^{I} = & \mathbf R^{I} + \mathbf \omega_u \odot pool(\mathbf R^{I}),
		\end{aligned}
	\end{equation}
	The CIN is utilized in pure id-relevant feature of CFD to enhance the instance-specific representations and filter out id-irrelevant information (e.g., illumination, background, viewpoints), towards better discriminative id-relevant information. 
	Fig.~\ref{fig:norm} shows the details of the calibrated-and-standardized batch normalization and the calibrated instance normalization.

	\subsection{Overall Loss Function}
	To adequately learn discriminative feature representations, we propose to adaptive distill identity-relevant feature from the entangled feature and restitute it to the network to ensure high discrimination. Specifically, we add the ReID loss on the discriminative identity-relevant feature $\mathbf R^{I} $. Meanwhile, we also add a domain loss into the contributive domain-relevant features $\mathbf R^{D}$ from the 4 ResNet blocks to ensure that id-irrelevant domain information does not flow into the subsequent network. The overall loss is as
	\begin{equation}
		\mathcal L_{Total} = \mathcal{L}_{ReID}(\mathbf{ R^{I}}) + \sum_{i=1}^{4}{\lambda_{i} \cdot \mathcal{L}_{Domain}(\mathbf{ R^{D}_{i}})}.
	\end{equation}
	where $\mathcal{L}_{ReID}$ denotes the triplet loss and ID loss, $\mathcal{L}_{Domain}$ denotes domain classification loss, and $\{\lambda_{i}\}|_{i=1}^4$ denote hyper-parameters for balancing the losses.  

	\section{Experiments}

	\begin{table*}[t]
		
		\caption{Comparison with state-of-the-art domain generalization methods on four large-scale person ReID benchmarks under Protocol-2 and Protocol-3 --- Market-1501 (M), Cuhk-SYSU (CS), CUHK03 (C3) and MSMT17 (MT). The performance is evaluated quantitatively by mean average precision (mAP) and cumulative matching characteristic (CMC) at Rank-1 (R1).}
		\vspace{-.15in}
		\centering
		\label{tab:sota}
		\fontsize{7.6pt}{5.6pt}\selectfont
		\begin{threeparttable}
			\begin{tabular}{p{3.0cm}|p{1.59cm}<{\centering}|p{0.9cm}<{\centering}|p{0.9cm}<{\centering}|p{0.8cm}<{\centering}p{0.8cm}<{\centering}|p{1.5cm}<{\centering}|p{0.9cm}<{\centering}|p{0.9cm}<{\centering}|p{0.8cm}<{\centering}p{0.8cm}<{\centering}}
				\toprule
				\multirow{2}{*}{Method} & \multirow{2}{*}{Source} & \multirow{2}{*}{IDs} & \multirow{2}{*}{Images} & \multicolumn{2}{c|}{Market-1501}& \multirow{2}{*}{Source} & \multirow{2}{*}{IDs} & \multirow{2}{*}{Images} & \multicolumn{2}{c}{DukeMTMC} \\
				&&&& mAP & R1 &&&& mAP & R1\\
				\midrule
				SNR~\cite{jin2020style}& \multirow{2}{*}{Com-MT} & \multirow{2}{*}{4,101} & \multirow{2}{*}{126,441} &41.4 & 70.1 & \multirow{2}{*}{Com-MT} & \multirow{2}{*}{4,101} & \multirow{2}{*}{126,441} & 50.0 & 69.2\\
				QAConv$_{50}$~\cite{liao2020interpretable} &  &  &  & 43.1 & 72.6 &  &  &  & 52.6 & 69.4 \\
				
				
				\midrule
				QAConv$_{50}$~\cite{liao2020interpretable}* & \multirow{5}{*}{\shortstack{MT+D+C3}} &\multirow{5}{*}{2,510} & \multirow{5}{*}{56,508} & 39.5 & 68.6 & \multirow{5}{*}{\shortstack{MT+M+C3}}& \multirow{5}{*}{2,559} & \multirow{5}{*}{52,922} & 43.4 & 64.9 \\
				SNR~\cite{jin2020style}* &  &  &  & 38.0 & 69.7 &  &  &  & 36.0 & 56.5   \\
				RDSBN~\cite{Bai_2021_CVPR} &  &  &   & 40.3 & 67.5 &  &  &  & - & - \\
				M$^3$L~(ResNet-50)~\cite{zhao2021learning} &  &  &   & 51.1 & 76.5 &  &  &  & 48.2 & 67.1 \\
				M$^3$L~(IBN-Net50)~\cite{zhao2021learning} &  &  &  &{52.5} & {78.3} &  &  &  & {48.8} &  {67.2} \\
				\midrule
				
				SNR~\cite{jin2020style}* & \multirow{6}{*}{\shortstack{MT+CS+C3}} &\multirow{6}{*}{13,742} & \multirow{6}{*}{72,190} & 34.6 & 62.7 & \multirow{6}{*}{\shortstack{MT+M+C3}}& \multirow{6}{*}{2,559} & \multirow{6}{*}{52,922} & - & - \\
				M$^3$L~(ResNet-50)~\cite{zhao2021learning}* &  &  &   & 58.4 & 79.9 &&  &  & - &  -  \\
				M$^3$L~(IBN-Net50)~\cite{zhao2021learning}* &  &  &   &  61.5&  82.3 &&  &  & - &   -  \\
				QAConv$_{50}$~\cite{zhao2021learning}* &  &  &   &  63.1&  83.7 &&  &  & - &   -  \\
				MetaBIN~\cite{choi2020meta}* &  &  &   &57.9  & 80.0  &&  &  & - &   -  \\
				CFD~(ResNet-50) &  &  &  & \textbf{70.0} &\textbf{87.2} &   &  &  & - &   -  \\
				
				\midrule
				
				SNR~\cite{jin2020style}*& \multirow{4}{*}{\shortstack{Com-\\(MT+D+C3)}} &\multirow{4}{*}{7,380} & \multirow{4}{*}{176,948} & 51.2 & 79.3  & \multirow{4}{*}{\shortstack{Com-\\(MT+M+C3)}}& \multirow{4}{*}{7,069} & \multirow{4}{*}{169,956} &  50.3 & 69.6  \\
				M$^3$L~(ResNet-50)~\cite{zhao2021learning} &  &  &   &  51.9 &  76.8 &  &  &  & 51.3 & 69.1  \\
				M$^3$L~(IBN-Net50)~\cite{zhao2021learning} &  &  &  & 57.2 &  80.2 & &  &  & 54.1 & 71.9  \\
				RaMoE~\cite{dai2021generalizable}  &  &  & & 56.5&82.0 &  &  &  & 56.9&73.6 \\
				\midrule
				
				SNR~\cite{jin2020style}* & \multirow{6}{*}{\shortstack{Com-\\(MT+CS+C3)}} &\multirow{6}{*}{17,502} & \multirow{6}{*}{175,111} & 52.4 & 77.8 & \multirow{6}{*}{\shortstack{Com-\\(MT+M+C3)}}& \multirow{6}{*}{7,069} & \multirow{6}{*}{169,956} & - &  -  \\
				M$^3$L~(ResNet-50)~\cite{zhao2021learning}* &  &  &   & 61.2 & 81.2 &  &  &  & - &  -   \\
				M$^3$L~(IBN-Net50)~\cite{zhao2021learning}* &  &  &   & 62.4 & 82.7 &  &  &  & - &  -  \\
				QAConv$_{50}$~\cite{zhao2021learning}* &  &  &   &  66.5&  85.0 &&  &  & - &   -  \\
				MetaBIN~\cite{choi2020meta}* &  &  &   & 67.2 & 84.5  &&  &  & - &   -  \\
				CFD~(ResNet-50) &  &  &  & \textbf{79.1} &\textbf{91.8} &   &  &  & - &  -     \\
				\bottomrule
				
				\toprule
				\multirow{2}{*}{Method} & \multirow{2}{*}{Source} & \multirow{2}{*}{IDs} & \multirow{2}{*}{Images} & \multicolumn{2}{c|}{CUHK03}& \multirow{2}{*}{Source} & \multirow{2}{*}{IDs} & \multirow{2}{*}{Images} & \multicolumn{2}{c}{MSMT17} \\ 
				&&&& mAP & R1 &&&& mAP & R1\\
				
				\midrule
				QAConv$_{50}$~\cite{liao2020interpretable} & Com-MT & 4,101 & 126,441 & 22.6 & 25.3 & D & 702 & 16,522 & 8.9 & 29.0 \\
				
				
				\midrule
				QAConv$_{50}$~\cite{liao2020interpretable}* & \multirow{4}{*}{MT+D+M} &\multirow{4}{*}{2,494} & \multirow{4}{*}{62,079} & 19.2 & 22.9 & \multirow{4}{*}{\shortstack{D+M+C3}}& \multirow{4}{*}{2,220} & \multirow{4}{*}{36,823} & 10.0 & 29.9 \\
				SNR~\cite{jin2020style}* &  &  &  &12.2 & 12.1&  &  &  & 9.3 & 27.0 \\
				M$^3$L~(ResNet-50)~\cite{zhao2021learning} &  & & & 30.9 & {31.9} &  &  &  & 13.1 & 32.0 \\
				M$^3$L~(IBN-Net50)~\cite{zhao2021learning} &  &  &  & {31.4} & 31.6 &  &  &  & {15.4} & {37.1} \\
				\midrule
				
				SNR~\cite{jin2020style}* & \multirow{6}{*}{\shortstack{MT+CS+M}} &\multirow{6}{*}{13,762} & \multirow{6}{*}{77,761} & 8.9 & 8.9 & \multirow{6}{*}{\shortstack{CS+M+C3}}& \multirow{6}{*}{13,452} & \multirow{6}{*}{54,878} & 6.8 & 19.9 \\
				M$^3$L~(ResNet-50)~\cite{zhao2021learning}* &  &  &   & 20.9 & 31.9 &&  & &15.9 & 36.9  \\
				M$^3$L~(IBN-Net50)~\cite{zhao2021learning}* &  &  &   & 34.2 & 34.4 &&  &  & 16.7& 37.5  \\
				QAConv$_{50}$~\cite{zhao2021learning}* &  &  &   &  25.4&  24.8 &  &  &  & 16.4 &  45.3  \\
				MetaBIN~\cite{choi2020meta}* &  &  &   & 28.8 & 28.1  & &  & &   17.8 &   40.2\\
				CFD~(ResNet-50) &  &  &  & \textbf{36.9} &\textbf{36.9} &   &  &  &\textbf{20.2} &\textbf{47.6}  \\
				
				\midrule
				SNR~\cite{jin2020style}* & \multirow{4}{*}{\shortstack{Com-\\(MT+D+M)}} &\multirow{4}{*}{7,414} & \multirow{4}{*}{192,271} & 26.2  & 26.8  & \multirow{4}{*}{\shortstack{Com-\\(D+M+C3)}}& \multirow{4}{*}{4,780} & \multirow{4}{*}{79,926} & 13.9 & 36.9   \\
				M$^3$L~(ResNet-50)~\cite{zhao2021learning} &  &  &   & 32.9 &  34.5 &  &  &  &15.2 &  36.2 \\
				M$^3$L~(IBN-Net50)~\cite{zhao2021learning} &  &  &  &34.4 &  35.0 &  &  &  & 17.2& 39.9  \\
				RaMoE~\cite{dai2021generalizable}  &  &  & &  33.5&34.6&  &  &  & 13.5&34.1 \\
				\midrule
				
				SNR~\cite{jin2020style}* & \multirow{6}{*}{\shortstack{Com-\\(MT+CS+M)}} &\multirow{6}{*}{17,536} & \multirow{6}{*}{190,434} & 17.5 & 17.1 & \multirow{6}{*}{\shortstack{Com-\\(CS+M+C3)}}& \multirow{6}{*}{14,902} & \multirow{6}{*}{78,089} & 7.7& 22.0 \\
				M$^3$L~(ResNet-50)~\cite{zhao2021learning}* &  &  &   & 32.3 & 33.8 &&  & &16.2& 36.9 \\
				M$^3$L~(IBN-Net50)~\cite{zhao2021learning}* &  &  &   & 35.7 & 36.5 &&  & & 17.4& 38.6\\
				QAConv$_{50}$~\cite{zhao2021learning}* &  &  &   &  32.9 &  33.3 &&  &  & 17.6 &  46.6  \\
				MetaBIN~\cite{choi2020meta}* &  &  &   & 43.0 & 43.1  &&  &  & 18.8 & 41.2  \\
				CFD~(ResNet-50) &  &  &  & \textbf{47.9} &\textbf{49.0} &   &  &  &\textbf{25.4} &\textbf{54.1}  \\
				\bottomrule
				
			\end{tabular}
			\begin{tablenotes}
				\scriptsize
				\item[\dag ] We re-implement this work based on the authors' code on Github with the same source datasets as us. 
			\end{tablenotes}
		\end{threeparttable}
		\vspace{-.15in}
	\end{table*}

	\subsection{Datasets and Evaluation Settings}
	Due to the CVPR2022 General Ethical Conduct, we have taken down DukeMTMC~\cite{zheng2017unlabeled} and adopted the new protocols based on the previous protocols~\cite{zhao2020learning,dai2021generalizable,song2019generalizable,jia2019frustratingly,tamura2019augmented} to evaluate the generalization ability of the model for person Re-ID.
	
	\textbf{Protocol-1:} Following the previous methods~\cite{song2019generalizable,jia2019frustratingly,tamura2019augmented}, we employ the existing Re-ID benchmarks to evaluate the Re-ID model's generalization ability. Specifically, the existing large-scale Re-ID datasets are viewed as multiple source domains, and the small-scale Re-ID datasets are used as unseen target domains. As shown in Tab.~\ref{table:Protocol}, seen domains include CUHK02~\cite{li2013locally}, CUHK03~\cite{li2014deepreid}, Market-1501~\cite{zheng2015scalable} and CUHK-SYSU~\cite{xiao2017joint}. Unseen domains contain VIPeR~\cite{gray2008viewpoint}, PRID~\cite{hirzer2011person}, GRID~\cite{loy2009multi} and iLIDS~\cite{zheng2009associating}. All training sets and testing sets in the seen domains are used for model training. The four small-scale Re-ID datasets are tested respectively, where the final performances are obtained by the average of 10 repeated random splits of testing sets.
	
	\textbf{Protocol-2 and Protocol-3:} Considering that the image quality of the small-scale Re-ID datasets is quite poor, the performances on these datasets can not precisely reflect the generalization ability of a model in real scenarios. The previous methods~\cite{zhao2020learning,dai2021generalizable} thus set two new kinds of protocols (\ieno, leave-one-out setting) for four large-scale Re-ID datasets. Specifically, four large-scale Re-ID datasets (Market-1501~\cite{zheng2015scalable}, CUHK-SYSU~\cite{xiao2017joint}, CUHK03~\cite{li2014deepreid} and MSMT17~\cite{wei2018person}) are divided into two parts: three datasets as the seen domains for training and the remaining one as the unseen domain for testing. In addition, for protocol-2, the test set of the seen domains also are adopted for training model, and protocol-3 only adopts the train set of seen domains for model training. 
	
	For simplicity, in the following sections, we denote Market1501 as M, CUHK02 as C2, CUHK03 as C3, MSMT17 as MT and CUHK-SYSU as CS. The three different evaluation settings are shown in Tab.~\ref{table:Protocol}.

	
	\subsection{Implementation Details}
	
	We adopt ResNet50~\cite{he2016deep} as our backbone. Following the previous method~\cite{luo2019strong}, the last residual layer’s stride size is set to 1. The generalized mean Poolineg (GeM)~\cite{radenovic2018fine} with a batch normalization layer is used after the backbone to obtain the Re-ID features. Images are resized to 384 $\times$ 128, and the training batch size of each domain is set to $64$, including 2 identities and 32 images per identity per domain. For data augmentation, we use random flipping, random cropping and color jittering. 
	We train the model for 60 epochs. The learning rate is initialized as $3.5\times10^{-4}$.

	\subsection{Comparison to state-of-the-art methods}

	\textbf{Comparison under the Protocol-1.} As shown in Tab.~\ref{table:result1}, the training domains under the previous Protocol-1 setting include the five multiple source datasets~(\ieno, M, D, C2, C3, and CS). However, due to ethical conduct, we must take away the DukeMTMC dataset from these four training domains since this dataset has been discredited by the creators, and adopt the remaining M, C2, C3, and CS as our training domains. Although we used one less dataset, our method still has the best average performance on these four small test datasets. Specifically, the mean performance of our CFD outperforms MetaBIN~\cite{choi2020meta} by 1.5\% R-1 accuracy and 4.0\% mAP, and DIR-ReID~\cite{zhang2021learning} 3.1\% R-1 accuracy and 7.3\% mAP, respectively. When using the 4 datasets as the training domains, our CFD outperforms the 2nd competitor by 5.5\% in terms of the R-1 mean performance accuracy.
	
	\textbf{Comparison under the Protocol-2 and the Protocol-3.} As shown in Tab.~\ref{tab:sota}, we compare the proposed CFD with MetaBIN~\cite{choi2020meta}, $QAConv_{50}$~\cite{liao2019interpretable}, $M^3L$~\cite{zhao2020learning} and RaMoE~\cite{dai2021generalizable} under Protocol-2 and Protocol-3. It is mentioned that we replace the DukeMTMC dataset with the CUHK-SYSU dataset, and re-implement some works based on the authors' codes with the same source datasets.
	The results show that CFD outperforms the performances of these methods by a large margin. Specifically, our method improves the second-best $QAConv_{50}$ by 3.5\% R1 accuracy and 6.9\% mAP on Market-1501 under the Protocol-2. And under the setting of Protocol-3, our CFD also improves the second-best $QAConv_{50}$ by 6.8\% and 12.6\% in terms of R1 accuracy and mAP, respectively. 
	The results demonstrate that the strong domain generalization of our CFD is attributed to calibrated feature decomposition with the guidance of calibrated normalization.
	
	\subsection{Ablation Study}
	
	
	\textbf{Effectiveness of components in CFD.} 
	To investigate the effectiveness of each component in CFD, we conduct a series of ablation studies.
	As shown in Tab.~\ref{tab:compo}, the performance of CFD without our designed normalization outperforms the baseline only by 3.17\% R1 accuracy and 3.70\% mAP. It indicates that vanilla attention mechanism can bring a slight improvement.
	Meanwhile, the model with CSBN/CSIN also can improve the performance by 8.17\%/10.21\% R1 accuracy and 6.54\%/7.90\% mAP respectively, which demonstrates the effectiveness of these two kinds of novel normalization technologies.
	Moreover, our proposed CDM with CIN and CSBN outperform the baseline by 11.91\%/16.00\% R1 accuracy and 6.54\%/7.90\% mAP, respectively. 
	These results demonstrate that CDM with CIN and CSBN together can improve, even more, proving to be mutually beneficial. It also indicates that the CIN and CSBN can effectively guide CDM to enable better decomposition, which is integrated to fully exploit the effective information for improving the model generalization. 
	

	\begin{table}[t]
		\centering
		\caption{Ablation studies on CDM, CSBN and CIN. Models are trained with the other three datasets except the Market dataset. }%
		\label{tab:compo}
		\fontsize{7.2pt}{7pt}\selectfont
		\vspace{-3mm}
		\begin{tabular}{p{1.3cm}|p{0.9cm}<{\centering}|p{0.9cm}<{\centering}|p{0.9cm}<{\centering}|p{0.9cm}<{\centering}p{0.9cm}<{\centering}}
			\toprule
			\multirow{2}{*}{Backbone} & \multirow{2}{*}{CDM} & \multirow{2}{*}{CIN} & \multirow{2}{*}{CSBN}& \multicolumn{2}{c}{MT+CS+C3$\rightarrow$M}\\ 
			&&&& mAP & R1 \\
			
			%
			\midrule
			\multirow{8 }{*}{ResNet-50} & Off& $\times$& $\times$ &  53.98 & 75.27 \\
			& Off& $\checkmark$& $\times$ &   64.19& 83.17 \\
			& Off& $\times$& $\checkmark$ &  62.15 & 81.81  \\
			& Off& $\checkmark$& $\checkmark$ &  66.63 &  84.78\\
			& On & $\times$& $\times$ &  57.68 &78.44  \\
			& On & \checkmark& $\times$ &   64.74 &82.96  \\
			& On & $\times$& \checkmark &   68.71 &86.43   \\
			& On & \checkmark& \checkmark &\textbf{69.98} & \textbf{87.18}   \\
			\bottomrule
		\end{tabular}
		\vspace{-3mm}
	\end{table}

	\begin{table}[tb]
		\caption{Effectiveness of study on different decomposition strategy in the ResNet architecture.}  
		
		\centering
		\fontsize{8pt}{8pt}\selectfont
		\vspace{-3mm}
		\begin{tabular}{c|cccc}
			\toprule
			\multirow{2}{*}{Method} & \multicolumn{4}{c}{MT+CS+C3$\rightarrow$M} \\ 
			
			& R1 & R5 & R10 & mAP  \\ 
			\midrule
			Baseline  &75.27   & 88.06    & 91.86     & 53.98  \\ 
			PFD  &77.27   & 89.96    & 93.61     & 56.18  \\ 
			CFD  &78.40   & 90.23    & 93.93     & 57.68  \\ 
			\midrule
			Baseline+CIN+CSBN  & 83.73 & 93.41 & 95.99  & 64.64 \\
			PFD+CIN+CSBN  &85.04&94.24&96.14&65.98 \\
			CFD+CIN+CSBN  &\textbf{87.18}  &\textbf{94.58}&\textbf{97.16}&\textbf{69.98} \\
			\bottomrule
		\end{tabular}
		\vspace{-0.60cm}
		\label{tab:ds}
	\end{table}

	\noindent\textbf{Influence of Disentanglement Design.}
	In our designed CFD module, as described in Subsection \ref{sec:cfd}, we use $\textbf a$ and $\textbf b$ as masks to extract pure domain-relevant feature $\mathbf R^- $, entangled feature $\mathbf R^{*}$ and pure id-relevant feature $\mathbf R^+ $ from the residual feature $\mathbf F$. Here, we study the influence of different disentanglement designs. 
	Following conventional disentanglement methods~\cite{zhang2021learning,jin2020style,zhang2021disentanglement,eom2019learning}, previous feature decomposition (PFD) methods directly disentangle the input feature $\mathbf F$ into id-relevant feature $\mathbf R^- $ and id-irrelevant/domain-related feature $\mathbf R^+ $. 
	Tab. \ref{tab:ds} shows the results. We observe that (1) without the designed normalization, our CFD outperforms PFD by \textbf{1.13\%} and \textbf{1.5\%} in Rank-1 and mAP for MT+CS+C3$\rightarrow$M, respectively; (2) with the designed normalization, our CFD outperforms PFD by \textbf{2.18\%/4.00\%} in Rank-1 and mAP on the unseen Market1501, respectively. The results demonstrate that the designed CFD is beneficial for the feature decomposition from the perspective of improving model generalization. Meanwhile, these two normalization technologies and CFD complement each other to improve the model's generalization.

	\section{Conclusion}
	In this paper, we propose a simple yet effective Calibrated Feature Decomposition module that focuses on improving the generalization capacity for person re-identification through a more judicious feature decomposition and reinforcement strategy. Specifically, we adopt the CSBN to learn calibrated person representation by jointly exploring intra-domain calibration and inter-domain standardization of multi-source domain features. CSBN restricts instance-level inconsistency of feature distribution for each domain and captures intrinsic domain-level specific statistics. Then, the above calibrated person representation is subtly decomposed into the identity-relevant feature, domain feature, and the remaining entangled one. For enhancing the generalization ability and ensuring high discrimination of identity-relevant features, a CIN is introduced to enforce discriminative id-relevant information and filter out id-irrelevant information, and meanwhile, the rich complementary clues from the remaining entangled feature are further employed to strengthen it. 
	
	\section{Limitations and broader impact}
	As for positive impacts, we demonstrate that through extensive experiments that a suitable domain generalizable method for person re-id contributes to performance on model generalization ability. We need not collect the images from the target domain and only use the existing data to train a robust model. Hence our domain generalization algorithm can have the potential to mitigate ethical concerns associated with the collecting of pedestrian data. 
	
	However, using a ReID system to identify pedestrians in a surveillance system may violate people's privacy. Because ReID systems typically (but not always) rely on unauthorized surveillance data, not all human subjects were aware they were being recorded. As a result, governments and officials must go to great lengths to create stringent regulations and legislation governing the use of ReID technology. 
	
	\appendix

	%
	%

	\section{Experiments}

	\subsection{Ablation Study for Where to Add Which Normalization Strategy.}

	\begin{table}[ht!]
		\centering
		\caption{Effectiveness of different normalization strategies in calibrated decomposition module. }
		\label{tab:dns}
		
		\begin{tabular}{p{2.0cm}<{\centering}|p{2.3cm}<{\centering}|p{0.9cm}<{\centering}p{0.9cm}<{\centering}}
			\toprule
			\multicolumn{2}{c|}{Calibrated Decomposition Module}& \multicolumn{2}{c}{MT+CS+C3$\rightarrow$M}\\ 
			Input Feature $\mathbf F$&pure ID Feature $\mathbf R^+$& mAP & R1 \\
			
			%
			\midrule
			-  & -   & 57.68 & 78.44 \\
			\midrule
			-  & IN  & 66.01 & 84.78 \\
			BN & -   & 58.08 & 79.08 \\
			BN & IN  & 65.51 & 84.00 \\
			\midrule
			-  & BN  & 59.00 & 78.47 \\ 
			IN & -   & 65.15 & 84.53\\ 
			IN & BN  & 65.15 &83.94 \\ 
			\midrule
			-    & CSBN & 67.99 & 85.21 \\
			CIN  & -    & 66.08 & 84.38 \\
			CIN  & CSBN & 69.00 & 86.97 \\ 
			CSBN & -    & 68.71 & 86.43 \\
			CSBN & CSBN & 65.29 & 84.12 \\ 
			\midrule
			-    & CIN  & 66.74 & 85.08 \\ 
			BN   & CIN  & 66.62 & 85.14 \\
			CBN  & CIN  & 69.27 & 86.89 \\ 
			CSBN & IBN  & 62.59 & 81.21  \\
			CSBN & IN   & 69.53 & 86.91 \\
			CSBN & CIN  & \textbf{69.98} & \textbf{87.18}   \\
			\bottomrule
		\end{tabular}
	\end{table}

	To investigate the effectiveness of different normalization strategies in calibrated decomposition module, we conduct a series of ablation studies. We mainly adopt different normalization strategy~(\ieno, BN, IN, CBN, IBN, CSBN, CIN) on the input $\mathbf F$ and pure ID Feature $\mathbf R^+$. Specifically, BN denotes to the batch normalization, IN denotes to the instance normalization, CBN refers to the calibrated BN~\cite{gao2021representative}, IBN refers to the ibn-net~\cite{luo2019strong}. CSBN and CIN are our designed normalization layer that is introduced in the Section 3.2 of manuscript paper.
	We observe that:
	
	(1) As shown in Tab.~\ref{tab:dns}, the performance of CFD without normalization only achieves 78.44\% R1 accuracy and 57.68\% mAP, respectively. It indicates that vanilla attention mechanism bring the slight improvement of model's generalizable ability. Meanwhile, the widely used BN also has a negligible impact on performance. 
	
	(2) As we know, IN could filter out some instance specific style information from the content to enhance model's generalizable ability. But the correct placement of IN is also not considered in the previous decoupling methods. As shown in Tab.~\ref{tab:dns}, the experiments illustrate that putting IN on the ID feature performs better than the input feature by about 1\%.
	The same conclusion is reached for the CSBN and CIN. These results demonstrate the importance of different types of normalization techniques placed in different positions on the corresponding features.
	
	(3) Thanks to the compensation of the identity-relevant information through the proposed CIN, our CIN achieves superior generalization capability, which outperforms other instance normalization technologies (IN and IBN). Our designed CSBN also shows the better generalization capability by exploring jointly intra-domain calibration and inter-domain standardization, which outperforms other batch normalization technologies (BN and CBN). It is worth noting that after we add the CSBN into the ID feature $\mathbf R^+$ for replacing the CIN, the performance dropped slightly. This result illustrates that the right normalization layer should be placed in the right place.
	
	(4) Moreover, the model with CSBN and CIN also can improve the performance by 8.74\% R1 accuracy and 12.30\% mAP respectively, which demonstrates the effectiveness of these two kinds of novel normalization technologies.
	These results demonstrate that CFD with CIN and CSBN together can improve, even more, proving to be mutually beneficial. It also indicates that the CIN and CSBN can effectively guide CFD to enable better decomposition, which is integrated to fully exploit the effective information for improving the model generalization.

	\subsection{Ablation Study for Where to Apply Calibrated Decomposition Module in the ResNet Architecture.}

	\begin{table}[!h]
		\caption{Effectiveness of study on where to apply Calibrated Decomposition Module in the ResNet architecture.}  
		\centering
		\begin{tabular}{c|cccc}
			\toprule
			\multirow{2}{*}{Method} & \multicolumn{4}{c}{CS+MT+C3$\rightarrow$M} \\ \cline{2-5}
			& R1 & R5 & R10 & mAP  \\ 
			\midrule
			Baseline &75.27 & 88.06  & 91.86 & 53.98 \\
			\midrule
			Stage-1  &81.68 & 92.40  & 95.25 & 59.78 \\
			Stage-2  &82.13 & 92.58  & 95.10 & 62.00 \\
			Stage-3  &83.58 & 93.26  & 95.64 & 63.05 \\
			Stage-4  &82.81 & 93.08  & 95.61 & 64.54  \\
			\midrule
			Stagesll    &\textbf{87.18}  &\textbf{94.58}&\textbf{97.16}&\textbf{69.98} \\
			\bottomrule
		\end{tabular}
		\label{tab:WTA}
	\end{table}

	We compare the cases of adding a single CFD module to a different convolutional block/stage, and to all the four stages (i.e., stage-1,2,3,4) of the ResNet-50 (see Tab.\ref{tab:WTA}). The module is added after the last layer of a convolutional block/stage. As shown in Tab.~\ref{tab:WTA}, in comparison with Baseline, the improvement from adding CFD is significant on stage-2, stage-3 and stage-4 and is a little smaller on stage-1. When CFD is added to all the four stages, we achieve the best performance.

	\subsection{Ablation Study for Domain Loss.}

	\begin{table}[!h]
		\caption{Effectiveness of domain loss on the CS+MT+C3$\rightarrow$M setting. ``B'' refers to the baseline model.}  
		\small
		\centering
		\begin{tabular}{l|cc}
			\toprule
			\multirow{2}{*}{Method} & \multicolumn{2}{c}{CS+MT+C3$\rightarrow$M} \\
			& R1  & mAP  \\ 
			\midrule
			B  &75.27        & 53.98  \\  
			B + Domain Loss &77.01   & 55.94  \\ 
			B + Domain Loss + CFD &78.44&57.68 \\ 
			B + CFD + CSBN + CIN  &{84.80} &{63.45} \\ 
			B + Domain Loss + CFD + CSBN + CIN  &\textbf{87.18} &\textbf{69.98} \\ 
			\bottomrule
		\end{tabular}
		\label{tab:DL}
	\end{table}

	As shown in Tab.~\ref{tab:DL}, we investigate the	influence of domain loss. We can observe that the Baseline model without domain loss obtain slight performance degradation over Baseline with domain loss. It demonstrates that the domain loss only brings the little improvement of model's generalizable ability.
	Meanwhile, we can observe that the CFD model without domain loss obtains performance degradation over CFD with domain loss. It indicates that domain loss can help the CFD to perform feature decomposition.

	\subsection{Ablation Study for Attention Choices.}
	
	\begin{table}[!h]
		\centering
		
		\caption{Ablation study for attention type choices for feature decomposition. ``S'' denotes spatial attention in which a spatial mask  of size ${1 \times H \times W}$ is learned. ``C'' denotes channel attention with a channel mask of size ${C \times 1 \times 1}$. ``SC'' denotes the joint using of both spatial attention and channel attention by multiplying them. We follow the implementations in CBAM \cite{woo2018cbam}.}
		\begin{center}
			\begin{tabular}{P{3.1cm}|C{0.8cm}C{0.8cm}C{0.8cm}C{0.8cm}}
				\toprule
				\multicolumn{1}{c|}{\multirow{2}{*}{Methods}} & \multicolumn{4}{c}{CS+MT+C3$\rightarrow$M}   \\
				\multicolumn{1}{c|}{}& mAP & R1 & mAP & R1  \\ 
				\midrule
				Baseline &75.27   & 88.06    & 91.86     & 53.98  \\ 
				Baseline+CFD w/ S &  81.21   &  91.39  &  94.03  &  58.16  \\ 
				Baseline+CFD w/ C&{87.18}  &{94.58}&{97.16}&{69.98} \\
				Baseline+CFD w/ SC &   88.87  & 95.67  & 97.62  & 73.11  \\
				\bottomrule
			\end{tabular}
		\end{center}
		\label{tab:ablation_attention}
	\end{table}

	We conduct an ablation study on the types of adopted attention. As experimental results shown in Tab.~\ref{tab:ablation_attention}, we find the SC is most effective for feature decomposition. While the spatial attention or channel-wise attention seem to be detrimental to the final performance. Although the best performance is based on the SC, we still adopt the channel attention in other experiment settings for fair comparisons. Because S is widely used on the disentanglement-based ReID task.

	\section{Visualization}
	
	\begin{figure*}[!h]
		\centerline{\includegraphics[width=1.0\linewidth]{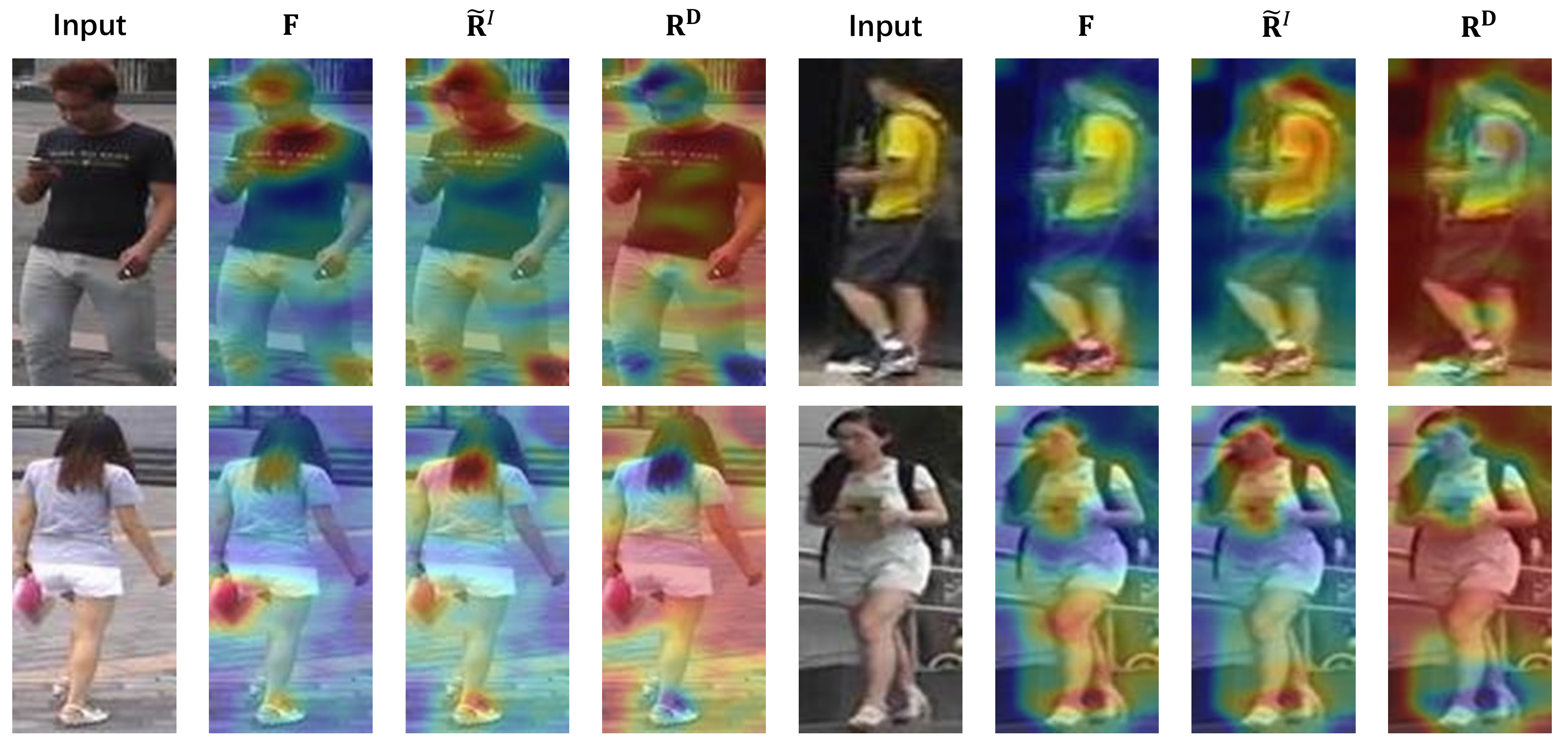}}
		
		\caption{Activation maps of different features of input feature $\mathbf F$, pure ID feature $\widetilde{\mathbf R}^I$, and domain feature $\mathbf R^D$ within a CFD module. They show that CFD can enhance the identity-relevant features (\ieno, pure ID feature $\widetilde{\mathbf R}^I$) well in the unseen target domain (\egno, Market1501), towards improving model's generalization ability. Meanwhile, There are still high responses of foreground on domain feature $\mathbf R^D$.These results illustrates that some crucial characteristics are stubbornly entwined in both the domain-relevant interference and identity-relevant feature, which are intractable to decompose in an unsupervised manner.}
		\label{fig:vis_ftp}
		\vspace{-5pt}
	\end{figure*}
	
	\subsection{Feature Map Visualization.} To better understand how a CFD module works, we visualize the intermediate feature maps of the fourth CFD module. Following \cite{zhou2019omni,zheng2021pose}, we get each activation map by summarizing the feature maps along channels following with the min-max normalization.
	
	
	Fig. \ref{fig:vis_ftp} shows the activation maps of input feature $\mathbf F$, pure ID feature $\widetilde{\mathbf R}^I$, and domain feature $\mathbf R^D$, respectively. 
	Although the input feature $\mathbf F$ mainly focus on foreground, there are still high response in the background and lack distinct salient responses.
	We can see that domain Feature $\mathbf R^D$ has high response simultaneously on background and foreground. It demonstrates that some crucial characteristics are stubbornly entwined in both the domain-relevant interference and identity-relevant feature, which are intractable to decompose in an unsupervised manner. In contrast, the pure ID Feature $\widetilde{\mathbf R}^I$ with the guidance of our designed normalization layers has high responses on salient regions of the human body, better capturing discriminative regions. 
	
	\begin{figure*}[!ht]
		\centerline{\includegraphics[width=1.0\linewidth]{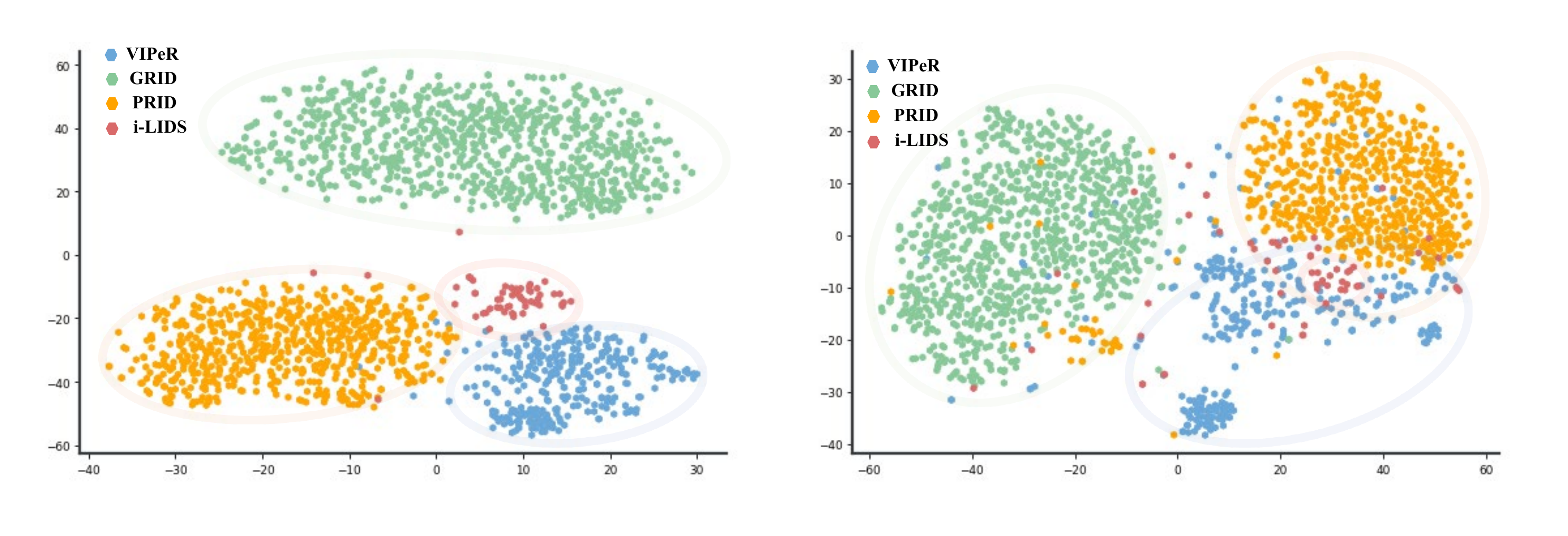}}
		
		\caption{The t-SNE visualization of the embedding vectors on four unseen target datasets (VIPeR, PRID, GRID, and i-LIDS). Best viewed in color and shape.}
		\label{fig:vis_ftp}
	\end{figure*}
	
	\subsection{Feature t-SNE Visualization.} We analyze our CFD framework and baseline through the performance and t-SNE visualization from the 4-th CFD module. In the baseline, the four unseen target domains are largely separately distributed and have an obvious domain gap. Thank to the CFD module, this domain gap has been eliminated. Meanwhile, our method shows that our method has shorter distances between different domains than those of baseline. Thus, we demonstrate that our CFD framework is generalizable and practical for real-world situations.
	
	{\small
		\bibliographystyle{ieee_fullname}
		\bibliography{egbib}
	}
	
\end{document}